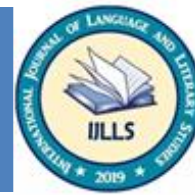

# Language Shift or Maintenance? An Intergenerational Study of the Tibetan Community in Saudi Arabia

**Sumaiyah Turkistani**
*King Saud University, English Department, College of Language Sciences*

**Mohammad Almoaily (corresponding author)**
*Associate Professor at the English Department, College of Language Sciences, King Saud University*
malmoaily@ksu.edu.sa

**Abstract**
*The present study provides the first-ever report on the language shift from Tibetan to Arabic among descendants of Tibetan families who migrated from the Tibet region to Saudi Arabia around 70 years ago. The aim of this study was to determine whether three age groups had adopted different practices in terms of maintaining Tibetan or shifting to Hijazi Arabic. To this end, 96 male and female members of the Tibetan community responded to a questionnaire in which they were asked about their code choice in different domains (home, neighbourhood, friends and relatives, expressing emotion, and performing religious rituals). The data revealed significant intergenerational differences between members of the community in terms of the extent of the shift to Arabic, with Tibetan rarely used by younger members and older members making only slightly more use of it. The difference between the three age groups was significant, at a p-value of .001.*

## 1. INTRODUCTION

When immigrants settle in a country whose language is foreign to them, they often find themselves in two conflicting scenarios: either they maintain their heritage language, which represents their culture and identity, or assimilate with the host community and shift to the new language (Tawalbeh et al., 2013). This study investigates language maintenance and shift in a speech community descending from Tibetan-speaking families who migrated to Saudi Arabia in the last 70 years. To do so, it uses narratives by elder members of the Tibetan community. It should be noted that the exact date of migration could not be determined due to the absence of official statistics and that the migration occurred in multiple waves.

To the best of our knowledge, this study provides the first-ever report of language shift and maintenance efforts in this speech community. The study aims to investigate the use of Arabic and Tibetan in various domains of language use, including the family, workplace, neighbourhood, and religion by male and female descendants of Tibetan immigrants belonging to three age groups (children, young adults, and older adults). By seeking to determine which language was most frequently used in various domains, we investigate the influence of age and gender on language shift. Thus, this study addresses the following research question:

1. Do younger generations of the Tibetan community in Saudi Arabia use Arabic more often than older generations?

Based on the literature review provided below, there is a higher expectation that older Tibetan speakers will maintain their heritage language compared to younger generations of the community. Thus, the following null and alternative hypotheses were formulated:

$H_0$: There is no difference between young and older generations in code choice between Arabic and Tibetan.

$H_a$: Younger Tibetan speakers use Arabic in their daily interactions more often than older generations.

## 2. LITERATURE REVIEW

### 2.1. Definition of Language Shift and Language Maintenance

Language maintenance and shift are two distinct phenomena, resulting mostly from prolonged contact between a minority group and a more dominant linguistic group. The term *language shift* was first used by Weinreich (1953) and later defined by Fishman (1989) as the process where speakers abandon their heritage language in favour of another language. Winford (2003) further clarified that language shift occurs when two distinct languages come into contact, resulting in either "partial or total abandonment of a group's native language in favor of another" (p. 15). This process can be driven by social, political, and economic pressures. In turn, it can lead to loss of linguistic and cultural heritage. This shift, according to Winford (2003), can be partial or full. In a partial language shift, individuals shift to the target language but do not fully acquire it. Conversely, a full language shift occurs when individuals achieve native-like proficiency and completely abandon their heritage language. According to Grenoble (2021), these two shift types are typically found in the speech of communities witnessing language shift across multiple generations: from a partial shift buttressed by bilingualism in the heritage and target languages in the first generation to a full shift to the target language in later generations. It must be noted that not every language shift necessarily leads to language death. Furthermore, not every language death is inevitably preceded by a language shift (Holmes & Wilson, 2017).

Mesthrie et al. (2009) define *language maintenance* as continuing to use and transmit a minority language from one generation to another, thereby enabling it to survive the various difficulties posed by the dominant language. Thus, language maintenance refers to the opposing scenario of language shift, as the minority group (or individuals from this group) successfully preserves their heritage language and continues to use it with full proficiency. This is normally achieved via deliberate steps to resist language shift (Fishman, 1989). Fishman (2001) further advanced that minority language maintenance is likely to occur through government, family, community, and neighbourhood support.

### 2.2. Inter-generational Language Maintenance and Shift

The phenomena of language maintenance and language shift have received attention from sociolinguists around the world. In this section, we review studies on different parts of the world regarding language shift and maintenance across different generations of speech communities.

Holmes and Wilson (2017) argued that the process of language shift is normally completed gradually by three to four generations of the immigrant family or minority speech community, whereby younger minority group speakers, who are mesmerized by the "glamor" of the dominant group's status, prestige, and social success, begin the process of abandoning their native language. This pattern has been found in several studies. For instance, Al-Khatib (2001) examined language shift and maintenance in the speech of Armenians living in Jordan and found that Arabic was deployed in most domains of language use. Furthermore, Armenian language use was very limited, mostly being used by the elder members of the community. In another study in the Middle Eastern context, Mugaddam (2006) examined language shift and maintenance among ethnic minority groups living in Sudan. The study investigated the participants' language proficiency, language use, and language attitudes. The results showed that although the sample had positive attitudes towards their ethnic languages, most of the younger generations had adopted Arabic as their primary language, using it in most contact domains. Veettil et al. (2020) also reported that second-generation immigrants of Malayalam speakers in Oman were shifting to Arabic despite efforts by the first generation of this speech community to maintain Malayalam, their heritage language. Stolberg's (2019) analysis of language shift from German to English among families of German origin in Canada also revealed that the shift occurred gradually between generations of the speech community, indicating that younger generations have led the shift process. Importantly, while children are more prone to language shifts, older members of a speech community are usually more inclined to maintain their heritage language (Verhaeghe et al., 2022; Zou, 2022).

Another sociolinguistic variable that influences the processes of language shift and maintenance is gender. Different cultural and societal norms may affect the specific gender leading the language shift process. For instance, if the women in a minority group are housewives and have little contact with the majority group, they are expected to maintain the heritage language. In other cases, young women with better job market opportunities are more likely to lead the shift to the majority language (Holmes & Wilson, 2017). Smith-Hefner (2009) reported that young Javanese show greater levels of shift to Indonesian patterns due to greater levels of engagement in communication and participation with the majority group. Kimani et al. (2018) showed that males and females show different shift patterns even at a young age. Young female students used English significantly more than young male students, who showed more maintenance of Sheng, their heritage language. However, contrary to the studies above, Abbasi et al. (2023) compared female and male university students' shift from Khowar to Urdu and English. The comparison revealed that female speakers maintained their heritage language more often than their male counterparts. These conflicting findings in the literature call for further investigations of the role of gender in language shift and maintenance, especially in under-researched speech communities.

### 2.3. Other Factors Influencing Language Shift and Maintenance

In addition to age and gender, other factors may influence the processes of language shift and maintenance, such as ethnicity, religion, socioeconomic status, family composition, community dynamics, and language proficiency. Attitudes towards heritage and majority languages can also be crucial factors in accelerating or hindering language shift. Dweik and Nofal (2013), for instance, reported that Indian ethnic minorities in Yemen could resist language shift and maintain their heritage languages because of attitudinal, linguistic, social, and political factors. Conversely, Hunt and Davis (2019) reported a number of factors that have accelerated the pace of the shift from German to English among families who migrated to Australia from the late 19th century, including assimilation with the majority group, anti-German sentiments, the relatively low number of the speech community's members who were not very well-connected, marriage to people outside of the community, and participation in the workforce. Giles et al. (1977) introduced the concept *ethnolinguistic vitality* to suggest that the pace of a language shift can be predicted by the presence or absence of linguistic, attitudinal, social, and cultural factors.

The present study has the potential to expand the conclusions drawn from previous research because of the ongoing nature of processes of language shift. Observing the process in real time in a community that has been neglected by researchers can provide valuable insights into this field of research.

## 3. METHODOLOGY

This study sought to determine whether there was a correlation between age and the extent to which Tibetan speakers in Saudi Arabia were shifting to Hijazi Arabic or maintaining their ethnic language (i.e. Tibetan). A mixed-methods research design was employed using a questionnaire to collect data from a sample of the Tibetan speech community. The study design was based on Dagamseh's PhD thesis, which drew on methods from earlier studies on language shift and maintenance by Al-Khatib (2001), Budiyana (2017), Dweik and Nofal (2013), Mugaddam (2006), Nesteruk (2010), Okpanachi and Joseph (2017), and Tawalbeh et al. (2013).

### 3.1. Sample

Purposive and snowball sampling were employed to recruit participants for the questionnaire. To achieve data homogeneity, all the participants polled in the current study lived in Taif, Saudi Arabia, at the time of their participation in the study. The study was confined to the city of Taif, the place of residence of most of the Tibetan community in Saudi Arabia. Therefore, the target population had better chances of maintaining their heritage language due to their ability to meet, socialize, and support one another.

A total of 100 participants received the questionnaire (see the questionnaire design below). The response rate 96%, meaning that 96 members volunteered to fill in the questionnaire. The participants were of three age groups: children (5–17), young adults (18–40), and older adults (40+). The young adults age group had the largest number of participants (n = 44). There were also 35 older adults 17 children. There were more females in the sample (n = 59) than males (n = 37). The members had varying levels of education, which was expected since they belonged to different age groups (see Table 1 below).

**Table 1**

*Demographic Information of the Questionnaire Sample*

| Demographic | Count | % |
|---|---|---|
| **Gender** | | |
| Male | 37 | 38.5% |
| Female | 59 | 61.5% |
| **Age** | | |
| 5–7 years | 17 | 17.7% |
| 8–40 years | 44 | 45.8% |
| > 40 years | 35 | 36.5% |
| **Marital status** | | |
| Single | 31 | 32.3% |
| Married | 57 | 59.4% |
| Divorced | 4 | 4.2% |
| Widow | 4 | 4.2% |
| **Last degree** | | |
| No degree | 3 | 3.1% |
| Elementary | 14 | 14.6% |
| Intermediate | 14 | 14.6% |
| Secondary | 26 | 27.1% |
| Bachelor's | 31 | 32.3% |
| Post-graduate degree | 8 | 8.2% |
| **Total Participants** | 96 | 100% |

None of the study participants had received formal education in Tibetan at the time of their participation in the study.

### 3.2. Questionnaire Design

First, a questionnaire was used to collect information about the participants (see Appendix A) regarding the language(s) they used in the various domains of language use. The questionnaire was divided into two main parts. In the first part, the respondents were prompted to answer demographic questions. In the second part, they were asked about the language they used when communicating with their family members, neighbours, friends and relatives, school and the workplace, in expressing emotions, and performing religious rituals. To help the participants report on their code choice in the various domains, a four-point categorical scale was used: *only Tibetan*, *only Arabic*, *both Tibetan and Arabic*, and *does not apply*.

### 3.3. Ethical considerations

The participants were informed of the voluntary nature of their participation in the study and that they had the freedom to withdraw at any time. They were also informed that their data were confidential and would only be used for study purposes and that their identities and personal information would not be shared with anyone. For the participants under 18, consent was also obtained from their parents. The research tool was approved by the Research Ethics Committee at KSU (approval no. KSU-HE-21-662).

## 4. RESULTS AND DISCUSSION
### 4.1. Domains of Tibetan and Arabic Language Use

In this section, we report on the language(s) used by the participants in the different domains. A comparison between the use of Tibetan and Arabic, especially among younger versus older members, would reveal whether the Tibetan community was shifting to Arabic or maintaining their heritage language. The data from the three age groups in the different domains were combined to test the hypotheses laid out in the introduction.

*Home Domain*

Table 2 identifies the languages used by the participants in the home environment with different family members. The data provide cross-generational information on the participants' language preference with other family members.

**Table 2**
*Language Use among Tibetan Family Members at Home*

| Family Members At Home Domain + Age Group | Only Tibetan | | Only Arabic | | Both Tibetan and Arabic | | Does Not Apply | |
|---|---|---|---|---|---|---|---|---|
| | n | % | n | % | n | % | n | % |
| Father | | | | | | | | |
| 5–17 | 0 | 0% | 17 | 100% | 0 | 0% | 0 | 0% |
| 18–40 | 4 | 9.09% | 27 | 61.36% | 11 | 25% | 2 | 4.54% |
| + 40 | 9 | 25.71% | 12 | 34.28% | 10 | 28.57% | 4 | 11.42% |
| Total: | 13 | 13.54% | 56 | 58.33% | 21 | 21.87% | 6 | 6.25% |
| Mother | | | | | | | | |
| 5–17 | 0 | 0% | 15 | 88.23% | 2 | 11.76% | 0 | 0% |
| 18–40 | 5 | 11.36% | 21 | 47.72% | 16 | 36.36% | 2 | 4.54% |
| + 40 | 11 | 31.42% | 10 | 28.57% | 12 | 34.28% | 2 | 5.71% |
| Total: | 16 | 16.66% | 46 | 47.91% | 30 | 31.25% | 4 | 4.16% |
| Brother | | | | | | | | |
| 5–17 | 0 | 0% | 13 | 76.47% | 0 | 0% | 4 | 23.52% |
| 18–40 | 1 | 2.27% | 39 | 88.63% | 2 | 4.54% | 2 | 4.54% |
| + 40 | 4 | 11.42% | 21 | 60% | 8 | 22.85% | 2 | 5.71% |
| Total: | 5 | 5.20% | 73 | 76.04% | 10 | 10.41% | 8 | 8.33% |
| **Sister** | | | | | | | | |
| 5–17 | 0 | 0% | 14 | 82.35% | 0 | 0% | 3 | 17.64% |
| 18–40 | 1 | 2.27% | 36 | 81.81% | 5 | 11.36% | 2 | 4.54% |
| + 40 | 3 | 8.57% | 15 | 42.85% | 14 | 40% | 3 | 8.57% |
| Total: | 4 | 4.16% | 65 | 67.70% | 19 | 19.79% | 8 | 8.33% |
| **Grandfather** | | | | | | | | |
| 5–17 | 2 | 11.76% | 15 | 88.23% | 0 | 0% | 0 | 0% |
| 18–40 | 8 | 18.18% | 17 | 38.63% | 7 | 15.90% | 12 | 27.27% |
| + 40 | 13 | 37.14% | 9 | 25.71% | 3 | 8.57% | 10 | 28.57% |
| Total: | 23 | 23.95% | 41 | 42.70% | 10 | 10.41% | 22 | 22.91% |
| **Grandmother** | | | | | | | | |
| 5–17 | 1 | 5.88% | 14 | 82.35% | 1 | 5.88% | 1 | 5.88% |
| 18–40 | 9 | 20.45% | 17 | 38.63% | 8 | 18.18% | 10 | 22.72% |
| + 40 | 16 | 45.71% | 8 | 22.85% | 3 | 8.57% | 8 | 22.85% |
| Total: | 26 | 27.08% | 39 | 40.63% | 12 | 12.5% | 19 | 19.79% |
| **Husband/Wife** | | | | | | | | |
| 5–17 | 0 | 0% | 0 | 0% | 0 | 0% | 17 | 100% |
| 18–40 | 1 | 2.27% | 31 | 70.45% | 11 | 25% | 1 | 2.27% |
| + 40 | 2 | 5.71% | 22 | 62.85% | 7 | 20% | 4 | 11.42% |
| Total: | 3 | 3.12% | 57 | 59.37% | 18 | 18.75% | 18 | 18.75% |
| **Son/Daughter** | | | | | | | | |
| 5–17 | 0 | 0% | 0 | 0% | 0 | 0% | 17 | 100% |
| 18–40 | 0 | 0% | 29 | 65.90% | 2 | 4.54% | 13 | 29.54% |
| + 40 | 4 | 11.42% | 26 | 74.28% | 4 | 11.42% | 1 | 2.85% |
| Total: | 4 | 4.1% | 59 | 61.45% | 6 | 6.25% | 27 | 28.12% |
| **Mean** | | | | | | | | |
| 5–17 | .375 | 2.20% | 11 | 64.69% | .375 | 2.20% | 5.25 | 30.88% |
| 18–40 | 3.625 | 8.23% | 27.125 | 61.64% | 7.75 | 15.11% | 5.5 | 12.5% |
| + 40 | 7.75 | 22.12% | 15.375 | 43.91% | 7.625 | 21.78% | 4.25 | 12.13% |

The figures in Table 1 above reveal that although Tibetan was not the preferred language of communication of any of the groups, interesting patterns could be observed. First, Tibetan was mostly used when speaking with older family members. Second, the older the participants, the more likely they were to use Tibetan to communicate with one another (compare the use of only Tibetan by older adults with their grandmothers (45%) with the use of only Tibetan by young participants with their brothers and sisters (0%). Third, Arabic was the children's preferred language to communicate with their parents. These are all indications that the Tibetan community is shifting to Arabic in the home domain.

*Neighbourhood Domain*

This section of the questionnaire aimed to report on the use of Tibetan with neighbours from the same speech community. Answers to this part of the questionnaire (see Table 3) helped us determine whether Tibetan had survived in out-of-home domains.

**Table 3**

*Tibetan Language Use with Tibetan Neighbours in the Neighbourhood*

| Neighbourhood Domain + Age Group | Only Tibetan | | Only Arabic | | Both Tibetan and Arabic | | Does Not Apply | |
|---|---|---|---|---|---|---|---|---|
| | n | % | n | % | n | % | n | % |
| 5–17 | 0 | 0% | 11 | 64.70% | 0 | 0% | 6 | 35.29% |
| 18–40 | 3 | 6.81% | 25 | 56.81% | 10 | 22.72% | 6 | 13.63% |
| + 40 | 9 | 25.71% | 14 | 40% | 9 | 25.71% | 3 | 8.57% |
| Total: | 12 | 12.5% | 50 | 52.08% | 19 | 19.79% | 15 | 15.62% |

None of the participants aged between five and 17 reported that they used Tibetan with other Tibetans in their neighbourhood. The young adults scored very low on using only Tibetan (6.81), while older adults had a slightly higher score on using only Tibetan to communicate with other members of the community (12.5%). These data also show that Tibetan was even less maintained in the neighbourhood domain, with younger members abandoning it altogether.

*Tibetan Friends and Relatives Domain*

In the *friends and relatives* domain, we identify whether the Tibetan community members maintained Tibetan in their outer circles.

**Table 4**

*Language Usage with Tibetan Friends and Relatives*

| Tibetan Friends Domain + Age Group | Only Tibetan | | Only Arabic | | Both Tibetan and Arabic | | Does Not Apply | |
|---|---|---|---|---|---|---|---|---|
| | n | % | n | % | n | % | n | % |
| 5–17 | 0 | 0% | 11 | 64.70% | 1 | 5.88% | 5 | 29.41% |
| 18–40 | 3 | 6.81% | 25 | 56.81% | 11 | 25% | 5 | 11.36% |
| + 40 | 10 | 28.57% | 18 | 51.42% | 6 | 17.14% | 1 | 2.85% |
| Total: | 13 | 13.54% | 54 | 56.25% | 18 | 18.75% | 11 | 11.45% |

As shown in Table 4, none of the child participants reported that they used Tibetan with their friends and relatives. The young adults rarely exclusively used Tibetan in this domain (6.8%), while the older adults used Tibetan only slightly more than those in the other age groups (28.5%).

*School and Workplace Domains*

School and the workplace constitute one of the most important domains of language use. Therefore, this part of the questionnaire investigated whether Tibetan was maintained in this domain or whether the participants were also shifting to Arabic when communicating with other members of the community at school or in the workplace.

**Table 5**
*Language Use in School and the Workplace with Tibetan Classmates and Colleagues*

| Workplace Domain Group | +Age | Only Tibetan | | Only Arabic | | Both Tibetan and Arabic | | Does Not Apply | |
|---|---|---|---|---|---|---|---|---|---|
| | | n | % | n | % | n | % | n | % |
| 5–17 | | 0 | 0% | 10 | 58.82% | 1 | 5.88% | 6 | 35.29% |
| 18–40 | | 0 | 0% | 34 | 77.27% | 8 | 18.18% | 2 | 4.54% |
| + 40 | | 4 | 11.42% | 19 | 54.28% | 6 | 17.14% | 6 | 17.14% |
| Total: | | 4 | 4.16% | 63 | 65.62% | 15 | 15.62% | 14 | 14.58% |

A similar pattern to the findings in the previous domains was found in this domain. Tibetan was not reported as the only language of communication between the children and young adults. Older adults, however, rarely used only Tibetan for communication (11.4%). This finding was not surprising because school and the workplace are the places where members of the community communicate the most with each other.

*Feelings and Emotions Domain*

Expressing feelings and emotions constitute a domain which reflects an individual's identity and fluency in a language. Thus, this part of the questionnaire sought to determine the language used by the participants when expressing their feelings and emotions and when performing their religious rituals.

**Table 6**

*Tibetan Language Use in Expressing Emotions and Practicing Religion*

| Expressing Feelings and Emotions Domain + Age Group | Only Tibetan | | Only Arabic | | Both Tibetan and Arabic | | Does Not Apply | |
|---|---|---|---|---|---|---|---|---|
| | **n** | **%** | **n** | **%** | **n** | **%** | **n** | **%** |
| Supplication | | | | | | | | |
| 5–17 | 0 | 0% | 17 | 100% | 0 | 0% | 0 | 0% |
| 18–40 | 0 | 0% | 44 | 100% | 0 | 0% | 0 | 0% |
| + 40 | 2 | 5.71% | 28 | 80% | 5 | 14.28% | 0 | 0% |
| Total: | 2 | 2.08% | 89 | 92.70% | 5 | 5.20% | 0 | 0% |
| Religious meetings in mosque | | | | | | | | |
| 5–17 | 0 | 0% | 17 | 100% | 0 | 0% | 0 | 0% |
| 18–40 | 0 | 0% | 42 | 95.45% | 2 | 4.54% | 0 | 0% |
| + 40 | 1 | 2.85% | 33 | 94.28% | 1 | 2.85% | 0 | 0% |
| Total: | 1 | 1.04% | 92 | 95.83% | 3 | 3.12% | 0 | 0% |
| Expressing anger | | | | | | | | |
| 5–17 | 0 | 0% | 17 | 100% | 0 | 0% | 0 | 0% |
| 18–40 | 3 | 6.81% | 35 | 79.54% | 6 | 13.63% | 0 | 0% |
| + 40 | 4 | 11.42% | 23 | 65.71% | 8 | 22.85% | 0 | 0% |
| Total: | 7 | 7.29% | 75 | 78.12% | 14 | 14.58% | 0 | 0% |
| Expressing excitement | | | | | | | | |
| 5–17 | 0 | 0% | 16 | 94.11% | 1 | 5.88% | 0 | 0% |
| 18–40 | 1 | 2.27% | 39 | 88.63% | 4 | 9.09% | 0 | 0% |
| + 40 | 4 | 11.42% | 26 | 74.28% | 5 | 14.28% | 0 | 0% |
| Total: | 5 | 5.20% | 81 | 84.37% | 10 | 10.41% | 0 | 0% |
| While asleep, language of dream | | | | | | | | |
| 5–17 | 0 | 0% | 17 | 100% | 0 | 0% | 0 | 0% |
| 18–40 | 1 | 2.27% | 40 | 90.90% | 3 | 6.81% | 0 | 0% |
| + 40 | 3 | 8.57% | 26 | 74.28% | 6 | 17.14% | 0 | 0% |
| Total: | 4 | 4.16% | 83 | 86.45% | 9 | 9.37% | 0 | 0% |
| **Mean** | | | | | | | | |
| 5–17 | 0 | 0% | 16.8 | 98.52% | .2 | 1.17% | 0 | 0% |
| 18–40 | 1 | 2.27% | 40 | 90.90% | 3 | 6.81% | 0 | 0% |
| + 40 | 2.8 | 7.99% | 27.2 | 77.71% | 5 | 14.28% | 0 | 0% |

The data in Table 6 above corroborate those in the previous domains. The child participants did not report using Tibetan as the only language for supplication, religious meetings, expressing emotions (anger and excitement), and dreaming. Younger adults rarely, if ever, used Tibetan as the sole language for expressing emotions or in religious rituals. Older adults, however, had slightly higher scores in these sub-domains. Furthermore, Arabic was clearly dominant among all the members of the sub-domains, with marginal inter-generational differences.

## 4.2. Tests of Significance

The alternative hypothesis in the current study predicted that younger members of the Tibetan community would use Arabic more frequently than older members. As shown in Tables 2 to 6 above, there was a clear trend in the data, which showed that the child participants used Arabic more frequently than Tibetan in the various domains included in the study. Another trend was that the use of Tibetan increased as the participants grew older. Table 7 below presents the findings of a one-way ANOVA analysis run to investigate the potential impact of age on the maintenance of Tibetan or the shift to Arabic in the domains included in the study.

**Table 7**

*One-way ANOVA Results: Effect of Age on Tibetan Language Use*

| Variables | Age | Mean | SD | Source | Sum of Squares | df | Mean Square | F | Sig. |
|---|---|---|---|---|---|---|---|---|---|
| Domain of language use | 5–7 | .70 | .25 | Between Groups | 2.150 | 2 | 1.075 | 9.270 | .001 |
| | 8–40 | .90 | .34 | Within Groups | 10.787 | 93 | .116 | | |
| | +40 | 1.12 | .37 | Total | 12.937 | 95 | | | |
| | Total | .95 | .37 | | | | | | |

The results of the one-way analysis of variance revealed a statistically significant difference between the members of the three age groups in terms of the use of Arabic and Tibetan in the various domains of language use (p-value = .001). Thus, the hypothesis that younger participants had shifted to Arabic at a higher rate than older the participants in the groups was substantiated.

The data revealed that the young participants showed a preference for the exclusive use of Arabic. Moreover, the young adults also revealed a high dependence on Arabic. Those in the older age group displayed higher rates of Tibetan language use, but Arabic was still dominant among them. This pattern of more frequent use of Arabic by the younger participants remained constant in all the domains of language use investigated in the current study. As detailed in the literature review section, this preference for the dominant language in interactions among younger people is conclusive in speech communities witnessing language shifts (see Al-Khatib, 2001; Mugaddam, 2006; Stolberg, 2019; Veettil et al., 2020). Therefore, from the findings of the current study, it is anticipated that future generations of the Tibetan community living in Saudi Arabia will shift completely to Arabic.

This witnessed shift to Hijazi Arabic in the data of the current study, especially among young participants, is in line with the abovementioned Giles' et al. (1977) ethnolinguistic vitality model, which suggests that that language shift is predicted by linguistic, attitudinal, social, and cultural factors. Based on the findings of this study, we shall argue that Tibetan in Saudi Arabic has a low ethnolinguistic vitality. Indeed, this shift towards Arabic seems to be driven by the presence of linguistic and social factors. First, Arabic is medium of instruction in Saudi public schools. Thus, young members of the Tibetan community acquire Arabic naturally due to being exposed to it at home and at school. Arabic is also the most frequently heard language in Saudi media outlets. It is also the language of religion and the language representing the local identity. This is important for the members who obtained a Saudi nationality. Arabic is also an essential language for job seekers in Saudi Arabia. Young Tibetans who notice the lack of need for Tibetan in the above social aspects are more likely to be less motivated to use it. Similarly, parents are expected to ensure that their children master the language used in education, the media, and the job market.

## 5. CONCLUSION

The current study aimed to investigate the extent of the shift from Tibetan to Arabic and efforts to maintain the heritage language by the Tibetan speech community living in Taif, Saudi Arabia. Data from three generations of the Tibetan community were examined to determine the extent of the shift from Tibetan to Arabic across the study population. The data revealed that young participants used Arabic significantly more than older members in various domains of language use, showing a greater degree of shift to Arabic among younger members of the community. However, Arabic was still dominant among all the group members included in the current study.

The findings of the current study suggest that the Tibetan speech community is in the process of shifting completely to Arabic. To preserve their heritage language, members of this minority group are advised to use Tibetan more often in the home domain. Additionally, more steps towards raising awareness among younger members of the cultural and historical value of the heritage language should be made in order to change existing attitudes towards Tibetan from negative or indifferent to positive.

Future research on language shift in the Tibetan community should include observation as a data collection tool, as responses to questionnaires or even interviews might not accurately reflect the real extent of the maintenance of or shift from Tibetan in Saudi Arabia.

## *AUTHORS' BIOS*

*Sumaiyah Turkistani is a PhD candidate at the English Department, King Saud University and a Lecturer at Jeddah University, Saudi Arabia. Sumaiyah has an MA in in English Language and Linguistics from Umm Al-Qura Universiy, Saudi Arabia.*

*Mohammad Almoaily is Associate Professor of Applied Linguistics at the Department of English Language at the College of Language Sciences, King Saud University. His Research Interests are language variation and change, code-switching, language policy, politeness, and pidginisation and creolization.*